\documentclass[letterpaper, 10 pt, conference]{ieeeconf}  

\IEEEoverridecommandlockouts                              

\usepackage{times}
\usepackage{amsfonts}
\usepackage{amsmath}
\usepackage{bm}
\usepackage{soul}

\usepackage{algorithm}
\usepackage{algorithmic}

\makeatletter
\let\NAT@parse\undefined
\makeatother

\usepackage[numbers]{natbib}
\usepackage{multicol}
\usepackage{multirow}
\usepackage[bookmarks=true]{hyperref}
\usepackage{graphicx}
\usepackage{color}
\usepackage[it,small]{caption}
\usepackage{subcaption}
\usepackage{dsfont}
\newcommand{\argmax}{\operatorname{arg\,max}}

\newcommand{\ve}[1]{\mathbf{#1}}

\usepackage{array}

 \belowdisplayskip \abovedisplayskip
\renewcommand{\baselinestretch}{0.97}

\newenvironment{packed_enum}{
\begin{enumerate}
  \setlength{\itemsep}{0pt}
  \setlength{\parskip}{0pt}
  \setlength{\parsep}{0pt}
}
{\end{enumerate}}

 \newlength\savedwidth

\newlength{\sectionReduceTop}
\newlength{\sectionReduceBot}
\newlength{\subsectionReduceTop}
\newlength{\subsectionReduceBot}
\newlength{\abstractReduceTop}
\newlength{\abstractReduceBot}
\newlength{\captionReduceTop}
\newlength{\captionReduceBot}
\newlength{\subsubsectionReduceTop}
\newlength{\subsubsectionReduceBot}

\newlength{\horSkip}
\newlength{\verSkip}

\newlength{\figureHeight}
\newcommand{\todo}[1]{\textcolor{blue}{\textbf{#1}}}

\title{\LARGE \bf
Recurrent Neural Networks for Driver Activity Anticipation \\ via Sensory-Fusion Architecture
}
\author{Ashesh Jain$^{1,2}$, Avi Singh$^{1}$, Hema S Koppula$^{1,2}$, Shane Soh$^{2}$, and Ashutosh Saxena$^{1,3}$
\thanks{$^1$Cornell University, $^2$Stanford University, $^3$Brain Of Things Inc.}
\thanks{{\tt ashesh@cs.stanford.edu, avisingh@iitk.ac.in, hema@cs.stanford.edu, shanesoh@stanford.edu, asaxena@cs.stanford.edu}}%
}

\begin{document}

\maketitle
\maketitle

\IEEEpeerreviewmaketitle

\begin{abstract}

Anticipating the future actions of a human is a widely studied problem in robotics that requires spatio-temporal reasoning. In this work we propose a deep learning approach for anticipation in sensory-rich robotics applications. We introduce a sensory-fusion architecture which jointly learns to anticipate and fuse information from multiple sensory streams. Our architecture consists of Recurrent Neural Networks (RNNs) that use Long Short-Term Memory (LSTM) units to capture long temporal dependencies. We train our architecture in a sequence-to-sequence prediction manner, and it \textit{explicitly} learns to predict the future given only a partial temporal context. We further introduce a novel loss layer for anticipation which prevents over-fitting and encourages early anticipation.  We use our architecture to anticipate driving maneuvers several seconds before they happen on a natural driving data set of 1180 miles. The context for maneuver anticipation comes from multiple sensors installed on the vehicle. Our approach shows significant improvement over the state-of-the-art in maneuver anticipation by increasing the precision from 77.4\% to \textbf{90.5\%} and recall from 71.2\% to \textbf{87.4\%}.

\end{abstract}

\vspace{\sectionReduceTop}
\section{Introduction}
\vspace{\sectionReduceBot}

Anticipating the future actions of a human is an important perception task and has many applications in robotics. It has enabled robots to navigate in a social manner and perform collaborative tasks with humans while avoiding conflicts~\citep{Kuderer12,Ziebart09,Koppula13,Wang13}. In another application, anticipating driving maneuvers several seconds in advance~\citep{Jain15,Morris11,Vasudevan12,Tawari14b} enables assistive cars to alert drivers before they make a dangerous maneuver. Maneuver anticipation complements existing Advance Driver Assistance Systems (ADAS) by giving drivers more time to react to road situations and thereby can prevent many accidents~\citep{Rueda04}.

Activity anticipation is a challenging problem because it requires the prediction of future events from  a limited temporal context. It is different from \textit{activity recognition}~\citep{Wang13}, where the complete temporal context is available for prediction.  Furthermore, in sensory-rich robotics settings, the context for anticipation comes from multiple sensors. In such scenarios the end performance of the application largely depends on how the information from different sensors are fused. Previous works on anticipation~\citep{Kitani12,Koppula13,Kuderer12} usually deal with single-data modality and do not address anticipation for sensory-rich robotics applications. Additionally, they learn representations using shallow architectures~\citep{Jain15,Kitani12,Koppula13,Kuderer12} that cannot handle long temporal dependencies~\citep{Bengio11}. 

\begin{figure}
\centering	
\includegraphics[width=\linewidth]{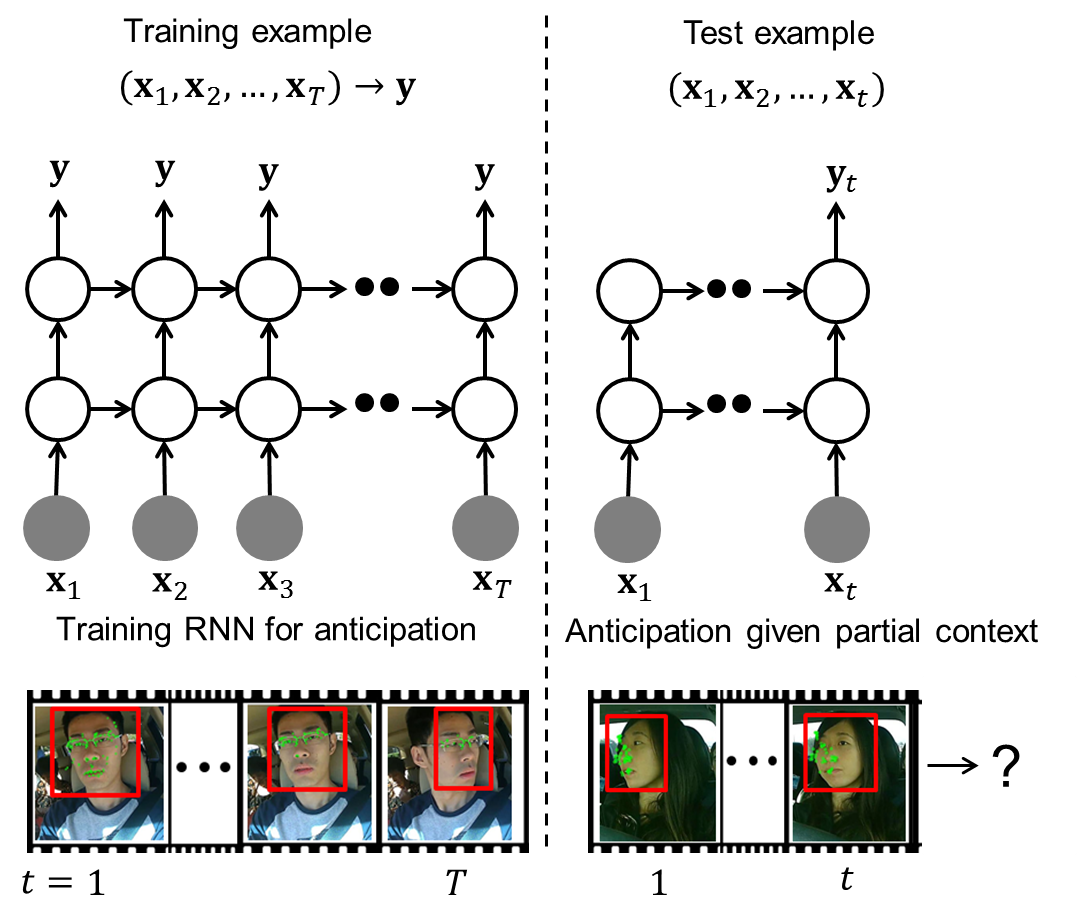}
\vspace{2.2\captionReduceTop}
\caption{(\textbf{Left}) Shows training RNN for anticipation in a sequence-to-sequence prediction manner. The network explicitly learns to map the partial context $(\ve{x}_1,..,\ve{x}_t)\;\forall t$ to the future event $\ve{y}$. (\textbf{Right}) At test time the network's goal is to anticipate the future event as soon as possible, i.e. by observing only a partial temporal context.}
\vspace{1.8\captionReduceBot}
\label{fig:introfig}
\end{figure}

In order to address the anticipation problem more generally, we propose a Recurrent Neural Network (RNN) based architecture which learns rich representations for anticipation. We focus on sensory-rich robotics applications, and our  architecture learns how to optimally fuse information from different sensors. Our approach  captures temporal dependencies by using Long Short-Term Memory (LSTM) units.  We train our architecture in a sequence-to-sequence prediction manner (Figure~\ref{fig:introfig}) such that it explicitly learns to anticipate given a partial context, and we introduce a novel loss layer which helps anticipation by preventing over-fitting.

 We evaluate our approach on the task of anticipating driving maneuvers several seconds before they happen~\citep{Jain15,Morris11}. The \textit{context} (contextual information) for maneuver anticipation comes from multiple sensors installed on the vehicle such as  cameras, GPS, vehicle dynamics, etc. Information from each of these sensory streams provides necessary cues for predicting future maneuvers. Our overall architecture models each sensory stream with an RNN and then non-linearly combines the high-level representations from multiple RNNs to make a final prediction.

\begin{figure}[t]
\centering	
\includegraphics[width=.9\linewidth]{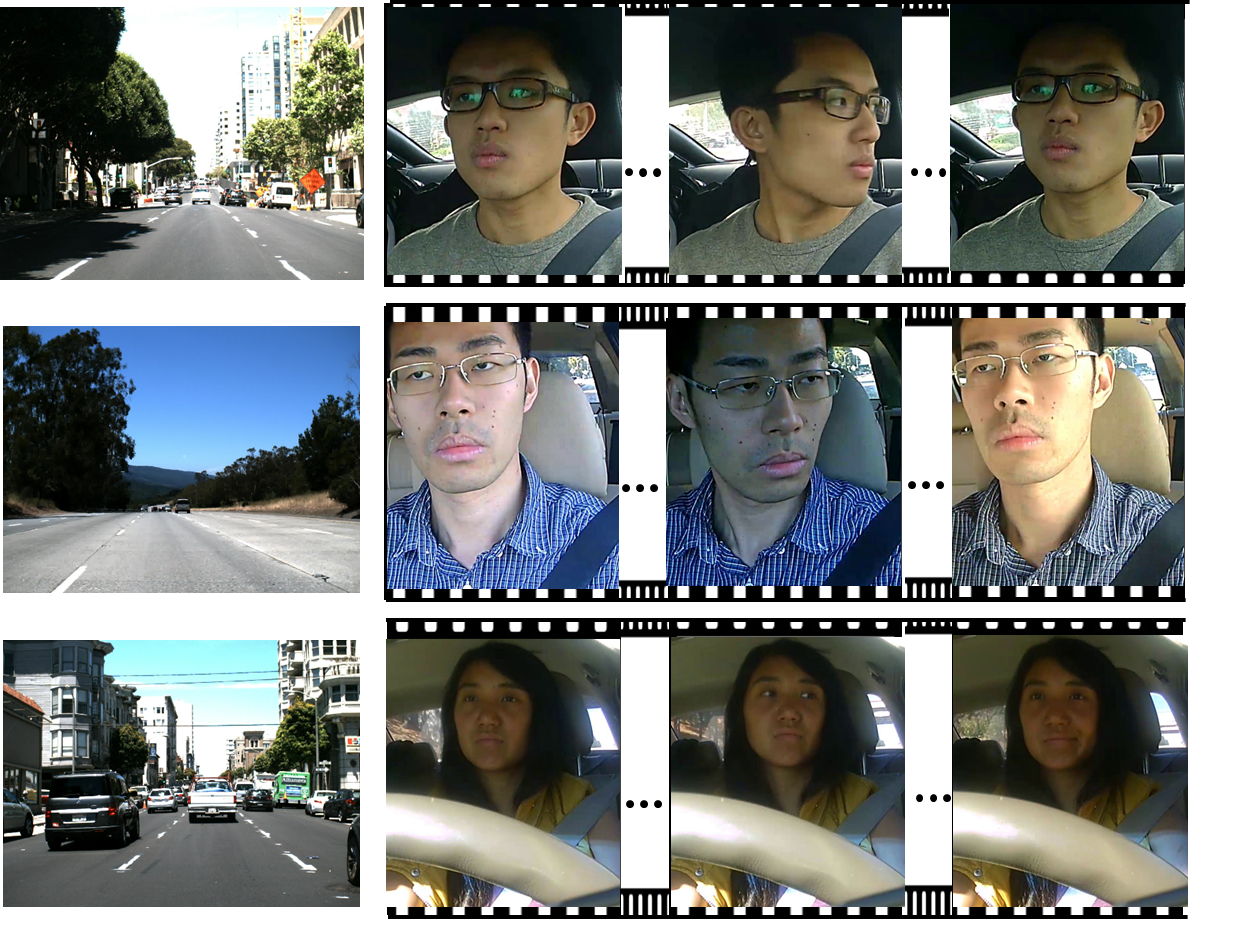}
\vspace{0.8\captionReduceTop}
\caption{\textbf{Variations in the data set.} {Images from the data set~\citep{Jain15} for} a left lane change. (\textbf{Left}) Views from the road facing camera. (\textbf{Right}) Driving style of the drivers vary for the same maneuver.}
\vspace{1.5\captionReduceBot}
\label{fig:dataset}
\end{figure}

 We report results on 1180 miles of natural driving data collected from 10 drivers~\citep{Jain15}. The data set is challenging because of the variations in routes and traffic conditions, and the driving styles of the drivers (Figure~\ref{fig:dataset}). On this data set, our deep learning approach improves the state-of-the-art in maneuver anticipation by increasing the precision from 77.4\% to \textbf{84.5\%} and recall from 71.2\% to \textbf{77.1\%}.  We further improved these results by extracting richer features from cameras such as the 3D head pose of the driver's face. Including these features into our architecture increases the precision and recall to \textbf{90.5\%} and \textbf{87.4\%} respectively. Key contributions of this paper are:
\begin{itemize}
\item A sensory-fusion RNN-LSTM architecture for anticipation in sensory-rich robotics applications.
\item A new vision pipeline with rich features (such as 3D head pose) for maneuver anticipation. 
\item State-of-the-art performance on maneuver anticipation on 1180 miles of driving data~\citep{Jain15}. 
\end{itemize}


\vspace{\sectionReduceTop}
\section{Related Work}
\vspace{\sectionReduceBot}

Our work is related to previous works on anticipating human activities, driver behavior understanding, and Recurrent Neural Networks (RNNs) for sequence prediction. 

Several works have studied human activity anticipation for human-robot collaboration and  forecasting. Anticipating human activities has been shown to improve human-robot collaboration~\citep{Wang13,Koppula13,Mainprice13,Dragan12}. Similarly, forecasting human navigation trajectories has enabled robots to plan sociable trajectories around humans~\citep{Kitani12,Bennewitz05,Kuderer12}. Feature matching techniques have been proposed for anticipating human activities from videos~\citep{Ryoo11}. Approaches used in these works learn shallow architectures~\citep{Bengio11} that do not properly model temporal aspects of human activities. Furthermore, they deal with a single data modality and do not tackle the challenges of sensory-fusion. We propose a deep learning approach for anticipation which efficiently handles temporal dependencies and learns to fuse multiple sensory streams.


We demonstrate our approach on anticipating driving maneuvers several seconds before they happen. This is a sensor-rich application for alerting drivers several seconds before they make a dangerous maneuvering decision. Previous works have addressed maneuver anticipation~\citep{BoschURBAN,Jain15,Morris11,Doshi11,Trivedi07} through sensory-fusion from multiple cameras, GPS, and vehicle dynamics. 
In particular, Morris et al.~\citep{Morris11} and Trivedi et al.~\citep{Trivedi07} used a Relevance Vector Machine (RVM) for intent prediction and performed sensory fusion by concatenating feature vectors. 


{More recently, Jain et al.~\citep{Jain15} showed that concatenation of sensory streams does not capture the rich context for modeling maneuvers. They proposed an Autoregressive Input-Output Hidden Markov Model (AIO-HMM) which fuses sensory streams through a linear transformation of features and it performs better than feature concatenation~\citep{Morris11}.} In contrast, we learn an expressive architecture to combine information from multiple sensors. Our RNN-LSTM based sensory-fusion architecture captures long temporal dependencies through its memory cell and learns rich representations for anticipation through a hierarchy of non-linear transformations of input data.  {Our work is also 
related to works on driver} behavior prediction with different sensors~\citep{Jabon10,Fletcher05,Fletcher03,Doshi11}, and vehicular controllers which act on these predictions~\citep{Shia14,Vasudevan12,Driggs15}.

Two building blocks of our architecture are Recurrent Neural Networks (RNNs)~\citep{Pascanu12} and Long Short-Term Memory (LSTM) units~\citep{Hochreiter97}. Our work draws upon ideas from previous works on RNNs and LSTM from the language~\citep{Sutskever14}, speech~\citep{Hannun14}, and vision~\citep{Donahue15} communities. 
Our approach to the joint training of multiple RNNs is related to the recent work on hierarchical RNNs~\citep{Du15}. We consider RNNs in multi-modal setting, which is related to the recent use of RNNs in image-captioning~\citep{Donahue15}. Our contribution lies in formulating activity anticipation in a deep learning framework using RNNs with LSTM units. We focus on sensory-rich robotics applications, and our architecture extends previous works on sensory-fusion from feed-forward networks~\citep{Ngiam11,Sung15} to the fusion of temporal streams. Using our architecture we demonstrate state-of-the-art results on maneuver anticipation.


\vspace{\sectionReduceTop}
\section{Preliminaries}
\vspace{\sectionReduceBot}

We now formally define anticipation and then present our Recurrent Neural
Network architecture. The goal of anticipation is to predict an event several
seconds before it happens given the contextual information up to the present
time. The future event can be one of multiple possibilities. At training
time a set of temporal sequences of {observations} and events
$\{(\mathbf{x}_1,\mathbf{x}_2,...,\mathbf{x}_T)_j,\ve{y}_j\}_{j=1}^N$ is provided
where $\mathbf{x}_t$ is the {observation} at time $t$, $\ve{y}$ is the representation of the event (described below) that happens at the end of the sequence  at $t= T$,
and $j$ is the sequence index. At test
time, however, the algorithm receives an {observation} $\mathbf{x}_t$ at each time step,
and its goal is to predict the future event as early as possible, i.e. by
observing only a partial sequence of {observations} $\{(\mathbf{x}_1,...,\mathbf{x}_t)
| t < T\}$. This differentiates anticipation from \textit{activity recognition}~\citep{Wang13b,Koppula13b} where in the latter  the complete {observation} sequence is available at  test time. In this paper, $\ve{x}_t$ is a real-valued feature vector and $\ve{y} =[y^1,...,y^K]$ is a vector of size $K$ (the number of events), where $y^k$ denotes the probability of the temporal sequence belonging to event the $k$ such that $\sum_{k=1}^K y^k = 1$. At the time of training, $\ve{y}$ takes the form of a one-hot vector with the entry in $\ve{y}$ corresponding to the ground truth event as $1$ and the rest $0$.

In this work we propose a deep RNN architecture with Long Short-Term Memory
(LSTM) units~\citep{Hochreiter97} for anticipation. Below we give an overview of
the standard RNN and LSTM which form the building blocks of our architecture
described in Section~\ref{sec:network}.

\vspace{\subsectionReduceTop}
\subsection{Recurrent Neural Networks}
\vspace{\subsectionReduceBot}

A standard RNN~\citep{Pascanu12} takes in a temporal sequence of vectors $(\mathbf{x}_1,\mathbf{x}_2,...,\mathbf{x}_T)$ as input, and outputs a sequence of vectors $(\mathbf{h}_1,\mathbf{h}_2,...,\mathbf{h}_T)$ also known as high-level representations. The representations are generated by non-linear transformation of the input sequence from $t=1	$ to $T$, as described in the equations below.
\begin{align}
\label{eq:h-rnn} \mathbf{h}_t &= f(\mathbf{W}\mathbf{x}_t + \mathbf{H}\mathbf{h}_{t-1} + \mathbf{b})\\
\label{eq:y-rnn} \mathbf{y}_t &= \texttt{softmax} (\mathbf{W}_y \mathbf{h}_t + \mathbf{b}_y)
\end{align}  
where $f$ is a non-linear function applied element-wise, and $\mathbf{y}_t$ is the \texttt{softmax} probabilities of the events having seen the {observations} up to $\mathbf{x}_t$.  $\mathbf{W}$, $\mathbf{H}$,  $\mathbf{b}$, $\mathbf{W}_y$,  $\mathbf{b}_y$ are the parameters that are learned. Matrices are denoted with bold, capital letters, and vectors are denoted with bold, lower-case letters.  In a standard RNN a common choice for $f$ is \texttt{tanh} or \texttt{sigmoid}. RNNs with this choice of $f$ suffer from a well-studied problem of \textit{vanishing gradients}~\citep{Pascanu12}, and hence are poor at capturing long temporal dependencies which are essential for anticipation. A common remedy to vanishing gradients is to replace \texttt{tanh} non-linearities by Long Short-Term Memory cells~\citep{Hochreiter97}.  We now give an overview of LSTM and then describe our model for anticipation. 

\vspace{\subsectionReduceTop}
\subsection{Long-Short Term Memory Cells}
\vspace{\subsectionReduceBot}

LSTM is a network of neurons that implements a memory cell~\citep{Hochreiter97}. The central idea behind LSTM is that the memory cell can maintain its state over time. When combined with RNN, LSTM units allow the recurrent network to remember long term context dependencies.

\begin{figure}[t]
\centering	
\includegraphics[width=.8\linewidth]{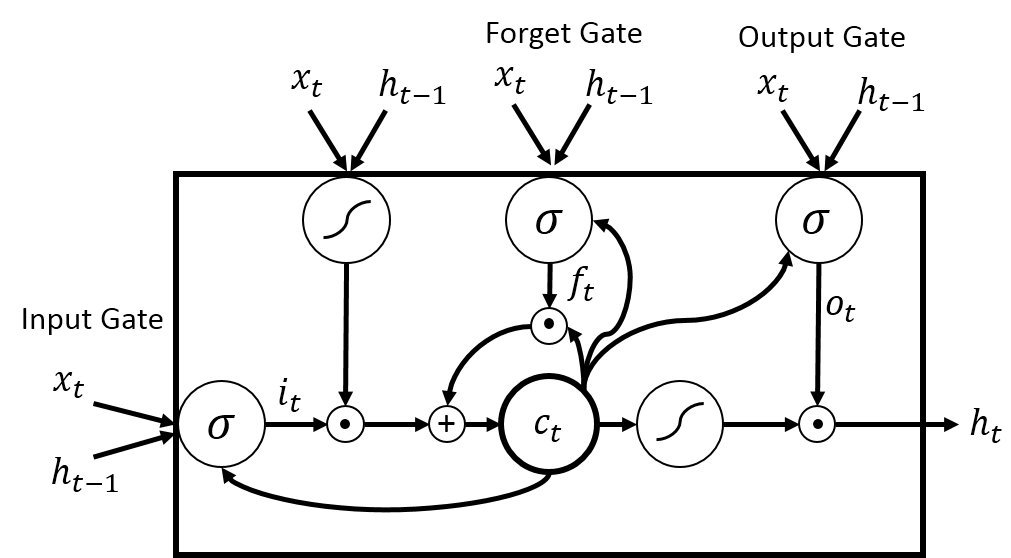}
\vspace{0.7\captionReduceTop}
\caption{\textbf{ Internal working of an LSTM unit.}}
\vspace{1.5\captionReduceBot}
\label{fig:lstm}
\end{figure}

LSTM consists of three gates -- input gate $\mathbf{i}$, output gate $\mathbf{o}$, and forget gate $\mathbf{f}$ -- and a memory cell $\mathbf{c}$. {See Figure~\ref{fig:lstm} for an illustration.} 
At each time step $t$, LSTM first computes its gates' activations \{$\mathbf{i}_t$,$\mathbf{f}_t$\}~\eqref{eq:i-lstm}\eqref{eq:f-lstm} and updates its memory cell from $\mathbf{c}_{t-1}$ to $\mathbf{c}_t$~\eqref{eq:c-lstm}, it then computes the output gate activation $\mathbf{o}_t$~\eqref{eq:o-lstm}, and finally outputs a hidden representation $\mathbf{h}_t$~\eqref{eq:h-lstm}. The inputs into LSTM are the {observations} $\mathbf{x}_t$ and the hidden representation from the previous time step $\mathbf{h}_{t-1}$. LSTM applies the following set of update operations:  
\begin{align}
\label{eq:i-lstm} \ve{i}_t &= \sigma(\ve{W}_{i}\ve{x}_t + \ve{U}_i \ve{h}_{t-1} + \ve{V}_i \ve{c}_{t-1} + \ve{b}_i)\\
\label{eq:f-lstm} \ve{f}_t &= \sigma(\ve{W}_{f}\ve{x}_t + \ve{U}_f \ve{h}_{t-1} + \ve{V}_f \ve{c}_{t-1} +\ve{b}_f)\\
\label{eq:c-lstm} \ve{c}_t &= \ve{f}_t \odot \ve{c}_{t-1} + \ve{i}_t \odot	 \texttt{tanh} (\ve{W}_{c}\ve{x}_t + \ve{U}_c \ve{h}_{t-1} +\ve{b}_c)\\
\label{eq:o-lstm} \ve{o}_t &= \sigma(\ve{W}_{o}\ve{x}_t + \ve{U}_o \ve{h}_{t-1} + \ve{V}_o \ve{c}_{t} +\ve{b}_o)\\
\label{eq:h-lstm} \ve{h}_t &= \ve{o}_t \odot \texttt{tanh}(\ve{c}_{t}) 
\end{align}

where $\odot$ is an element-wise product and $\sigma$ is the logistic function. $\sigma$ and \texttt{tanh} are applied element-wise. $\ve{W}_*$, $\ve{V}_*$, $\ve{U}_*$, and $\ve{b}_*$ are the parameters. The input and forget gates of LSTM participate in updating the memory cell~\eqref{eq:c-lstm}. More specifically, forget gate controls the part of memory to forget, and the input gate computes new values based on the current {observation} that are written to the memory cell. The output gate together with the memory cell computes the hidden representation~\eqref{eq:h-lstm}. Since  LSTM cell activation involves \textit{summation} over time~\eqref{eq:c-lstm} and derivatives distribute over sums, the gradient in LSTM gets propagated over a longer time before vanishing. In the standard RNN, we replace the non-linear $f$ in equation~\eqref{eq:h-rnn} by the LSTM equations given above in order to capture long temporal dependencies. We use the following shorthand notation to denote the recurrent LSTM operation.
\begin{equation}
(\ve{h}_t,\ve{c}_t) = \text{LSTM}(\ve{x}_t,\ve{h}_{t-1},\ve{c}_{t-1})
\end{equation}

We now describe our RNN architecture with LSTM units for anticipation. Following which we will describe a particular instantiation of our architecture for maneuver anticipation where the {observations} $\ve{x}$ come from multiple sources.

\vspace{\sectionReduceTop}
\section{Network Architecture for Anticipation}
\vspace{\sectionReduceBot}
\label{sec:network}
In order to anticipate, an algorithm must learn to predict the future {given only} a partial temporal context. This makes anticipation challenging and also differentiates it from activity recognition. Previous works treat anticipation as a recognition problem~\citep{Koppula13,Morris11,Ryoo11} and train discriminative classifiers (such as SVM or CRF) on the complete temporal context. However, at test time {these} classifiers only observe  a partial temporal context and make predictions within a filtering framework. We model anticipation with a recurrent architecture which unfolds through time. This lets us train a single classifier that learns how to handle partial temporal context of varying lengths.

Furthermore, anticipation in robotics applications is challenging because the contextual information can come from multiple sensors with different data modalities. Examples include autonomous vehicles that reason from multiple sensors~\citep{Geiger12} or robots that jointly reason over perception and language instructions~\citep{Misra14}. In such applications the way information from different sensors is fused is critical to the application's final performance. For example Jain et al.~\citep{Jain15} showed that for maneuver anticipation, learning a simple transformation of the sensory streams works better than direct concatenation of those streams. We therefore build an end-to-end deep learning architecture which jointly learns to anticipate and fuse information from different sensors.   

\vspace{\subsectionReduceTop}
\subsection{RNN with LSTM units for anticipation}
\vspace{\subsectionReduceBot}

At the time of training, we observe the complete temporal {observation sequence} and the event  $\{(\mathbf{x}_1,\mathbf{x}_2,...,\mathbf{x}_T),\ve{y}\}$. Our goal is to train a network which predicts the future event given a partial temporal {observation sequence} $\{(\mathbf{x}_1,\mathbf{x}_2,...,\mathbf{x}_t) 	| t < T\}$. We do so by training an RNN in a sequence-to-sequence prediction manner. Given training examples $\{(\mathbf{x}_1,\mathbf{x}_2,...,\mathbf{x}_T)_j,\ve{y}_j\}_{j=1}^N$ we train an RNN with LSTM units to map the sequence of {observations} $(\mathbf{x}_1,\mathbf{x}_2,...,\mathbf{x}_T)$ to the sequence of events  $(\ve{y}_1,...,\ve{y}_T)$ such that $\ve{y}_{t} = \ve{y}, \forall t$, as shown in Fig.~\ref{fig:introfig}. Trained in this manner, our RNN will attempt to map all sequences of partial {observations} $(\mathbf{x}_1,\mathbf{x}_2,...,\mathbf{x}_t)$~$\forall t \leq~T$ to the future event $\ve{y}$. This way our model explicitly learns to anticipate. We additionally use LSTM units which prevents the gradients from vanishing and allows our model to capture long temporal dependencies in human activities.\footnote{Driving maneuvers can take up to 6 seconds and the value of T can go up to 150 with a camera frame rate of 25 fps.}


\vspace{\subsectionReduceTop}
\subsection{Fusion-RNN: Sensory fusion RNN for anticipation}
\vspace{\subsectionReduceBot}
\label{sec:sensor-fusion-rnn}

We now present an instantiation of our RNN architecture for fusing two sensory streams: $\{(\ve{x}_1,...,\ve{x}_T), \;(\ve{z}_1,...,\ve{z}_T)\}$. In Sections~\ref{sec:maneuver} and~\ref{experiments}, we use the fusion architecture for maneuver anticipation.

\begin{figure}[t]
\centering	
\includegraphics[width=.9\linewidth]{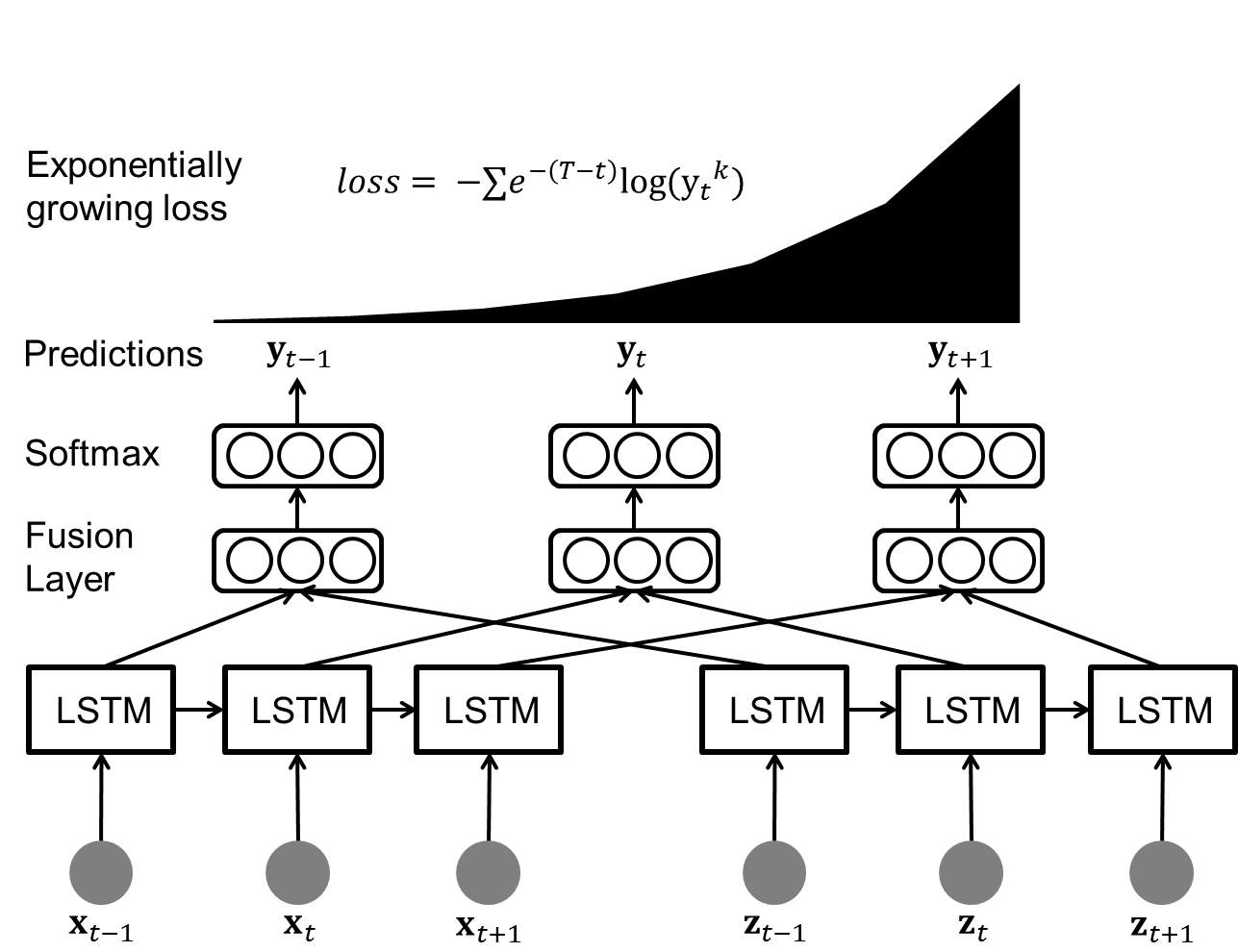}
\vspace{1.0\captionReduceTop}
\caption{\textbf{Sensory fusion RNN for anticipation.} (\textbf{Bottom}) In the Fusion-RNN each sensory stream is passed through their independent RNN. (\textbf{Middle}) High-level representations from RNNs are then combined through a fusion layer. (\textbf{Top}) In order to prevent over-fitting early in time the loss exponentially increases with time.}
\vspace{1.5\captionReduceBot}
\label{fig:fusion}
\end{figure}

An obvious way to allow sensory fusion in the RNN is by concatenating the streams, i.e. using $([\ve{x}_1;\ve{z}_1],...,[\ve{x}_T;\ve{z}_T])$ as input to the RNN. However, we found that this sort of simple concatenation performs poorly. We instead learn a sensory fusion layer which combines the high-level representations of sensor data. Our proposed architecture first passes the two sensory streams $\{(\ve{x}_1,...,\ve{x}_T), \;(\ve{z}_1,...,\ve{z}_T)\}$ independently through separate  RNNs~\eqref{eq:f-rnn1} and~\eqref{eq:f-rnn2}.  
The high level representations from both RNNs $\{(\ve{h}_1^x,...,\ve{h}_T^x), \;(\ve{h}_1^z,...,\ve{h}_T^z)$ are then concatenated at each time step $t$ and passed through a fully connected (fusion) layer which fuses the two representations~\eqref{eq:f-fusion}, {as shown in Figure~\ref{fig:fusion}}. The output representation from the fusion layer is then passed to the softmax layer for anticipation~\eqref{eq:f-output}. The following operations are performed from $t=1$ to $T$.
\begin{align}
\label{eq:f-rnn1} (\ve{h}_t^x,\ve{c}_t^x) &= \text{LSTM}_x(\ve{x}_t,\ve{h}_{t-1}^x,\ve{c}_{t-1}^x)\\
\label{eq:f-rnn2} (\ve{h}_t^z,\ve{c}_t^z) &= \text{LSTM}_z(\ve{z}_t,\ve{h}_{t-1}^z,\ve{c}_{t-1}^z)\\
\label{eq:f-fusion} \text{Sensory fusion:  } \ve{e}_t &= \texttt{tanh}(\ve{W}_f[\ve{h}_t^x;\ve{h}_t^z] + \ve{b}_f)\\
\label{eq:f-output} \ve{y}_t &= \texttt{softmax}(\ve{W}_y\ve{e}_t + \ve{b}_y)
\end{align}	 
where $\ve{W}_*$ and $\ve{b}_*$ are model parameters, and $\text{LSTM}_x$ and $\text{LSTM}_z$ process the sensory streams $(\ve{x}_1,...,\ve{x}_T)$ and $(\ve{z}_1,...,\ve{z}_T)$ respectively. {The same framework can be extended to handle more sensory streams.} 

\vspace{\subsectionReduceTop}
\subsection{Exponential loss-layer for anticipation.} 
\vspace{\subsectionReduceBot}
We propose a new loss layer which encourages the architecture to anticipate early while also ensuring that the architecture does not over-fit the training data early enough in time when there is not enough context for anticipation. 
When using the standard softmax loss, the architecture suffers a loss of $-\log(y_t^k)$ for the mistakes it makes at each time step, where $y_t^k$ is the probability of the ground truth event $k$ computed by the architecture using Eq.~\eqref{eq:f-output}. We propose to modify this loss by multiplying it with an exponential term as illustrated in Figure~\ref{fig:fusion}. Under this new scheme, the loss exponentially grows with time as shown below. 
\begin{align}
\label{eq:loss}
loss = \sum_{j=1}^N \sum_{t=1}^T  - e^{-(T-t)}\log (y_t^k)
\end{align}
This loss penalizes the RNN exponentially more for the mistakes it makes as it sees more {observations}. This encourages the model to fix mistakes as early as it can in time. The loss in equation~\ref{eq:loss} also penalizes the network less on mistakes made early in time when there is not enough context available. This way it acts like a regularizer and reduces the risk to over-fit very early in time. 

\vspace{\subsectionReduceTop}
\subsection{Model training and data augmentation}
\vspace{\subsectionReduceBot}
\label{subsec:augmentation}

Our architecture for maneuver anticipation has more than 25,000 parameters that need to be learned (Section~\ref{sec:maneuver}). With such a large number of parameters on a non-convex manifold, over-fitting becomes a major challenge. We therefore introduce redundancy in the training data which acts like a regularizer and reduces  over-fitting~\citep{Krizhevsky12,Hannun14}. 
 In order to augment training data, we extract sub-sequences of temporal {observations}. Given a training example with two temporal sensor streams $\{(\ve{x}_1,...,\ve{x}_T),(\ve{z}_1,...,\ve{z}_T),\ve{y}\}$, we uniformly randomly sample multiple sub-sequences $\{(\ve{x}_i,...,\ve{x}_j),(\ve{z}_i,...,\ve{z}_j),\ve{y} | 1 \leq i < j \leq T\}$ as additional training examples. It is important to note that data augmentation only adds redundancy and \textit{does not} rely on any external source of new information.

On the augmented data set, we train the network described in Section~\ref{sec:sensor-fusion-rnn}. We use RMSprop gradients which have been shown to work well on training deep networks~\citep{Dauphin15}, and we keep the step size fixed at $10^{-4}$. We experimented with different variants of softmax loss, and our proposed loss-layer with exponential growth Eq.~\eqref{eq:loss} works best for anticipation (see Section~\ref{experiments} for details).


\begin{figure*}[t]
\centering	
\includegraphics[width=.95\linewidth]{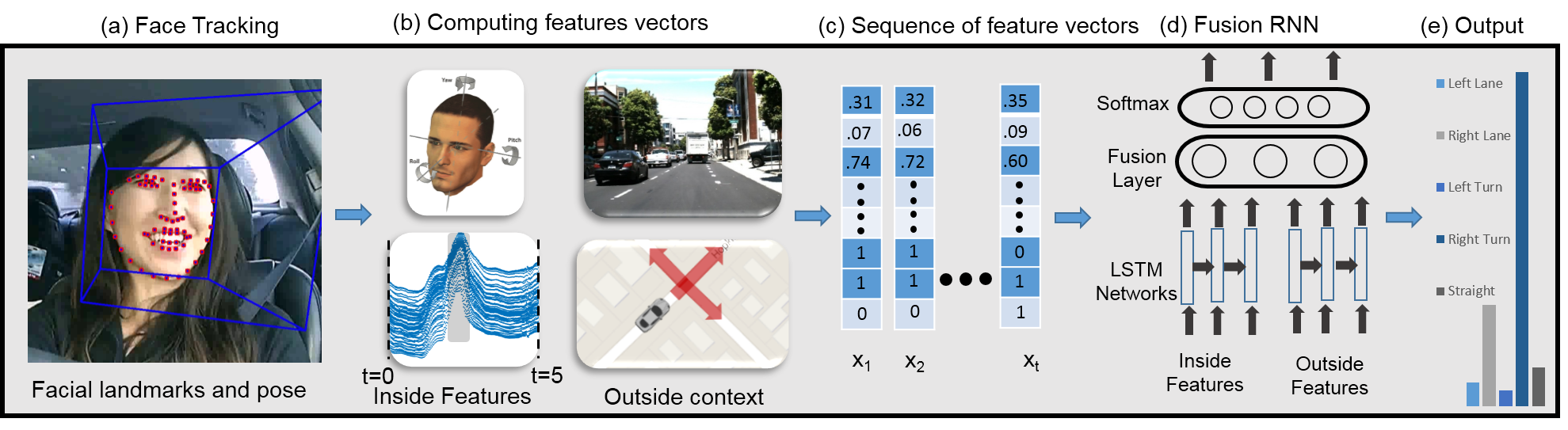}
\vspace{\captionReduceTop}
\caption{\textbf{Maneuver anticipation pipeline.} Temporal context in maneuver anticipation comes from cameras facing the driver and the road, the GPS, and the vehicle's dynamics. (\textbf{a}, \textbf{b} and \textbf{c}) We improve upon the vision pipeline from Jain et al.~\citep{Jain15} by tracking 68 landmark points on the driver's face and including the 3D head-pose features. (\textbf{d}) Using the Fusion-RNN we combine the sensory streams of features from inside the vehicle (driver facing camera), with the features from outside the vehicle (road camera, GPS, vehicle dynamics). {(\textbf{e}) The model outputs the predicted probabilities of future maneuvers.}}
\vspace{1.75\captionReduceBot}
\label{fig:sys_fig}
\end{figure*}

\vspace{\sectionReduceTop}
\section{Context for maneuver anticipation}
\vspace{\sectionReduceBot}
\label{sec:maneuver}
In maneuver anticipation the goal is to anticipate the driver's future maneuver several seconds before it happens~\citep{Jain15,Morris11}. The contextual information for anticipation is extracted from sensors installed in the vehicle. Previous work from Jain et al.~\citep{Jain15} considers the context from a driver facing camera, a camera facing the road in front, a Global Positioning System (GPS), and an equipment for recording the vehicle's dynamics. The overall contextual information from the sensors is grouped into: (i) the context from inside the vehicle, which comes from the driver facing camera and is represented as temporal sequence of features $(\ve{x}_1,...,\ve{x}_T)$; and (ii) the context from outside the vehicle, which comes from remaining sensors and is represented as $(\ve{z}_1,...,\ve{z}_T)$.

{We improve the pipeline from Jain et al.~\citep{Jain15} with our deep learning architecture and new features for maneuver anticipation. Figure~\ref{fig:sys_fig} shows our complete pipeline.} In order to anticipate maneuvers, our RNN architecture (Figure~\ref{fig:fusion}) processes the temporal context $\{(\ve{x}_1,...,\ve{x}_t),\; (\ve{z}_1,...,\ve{z}_t)\}$  at every time step $t$, and outputs softmax probabilities $\ve{y}_t$ for the following five maneuvers:  $\mathcal{M}=$~\{\textit{left turn}, \textit{right turn}, \textit{left lane change}, \textit{right lane change}, \textit{straight driving}\}.  We now give an overview of the feature representation used by Jain et al.~\citep{Jain15} and then describe our features which significantly improve the performance.

\begin{figure}[t]
\centering
\vspace{0.15in}
\includegraphics[width=.9\linewidth]{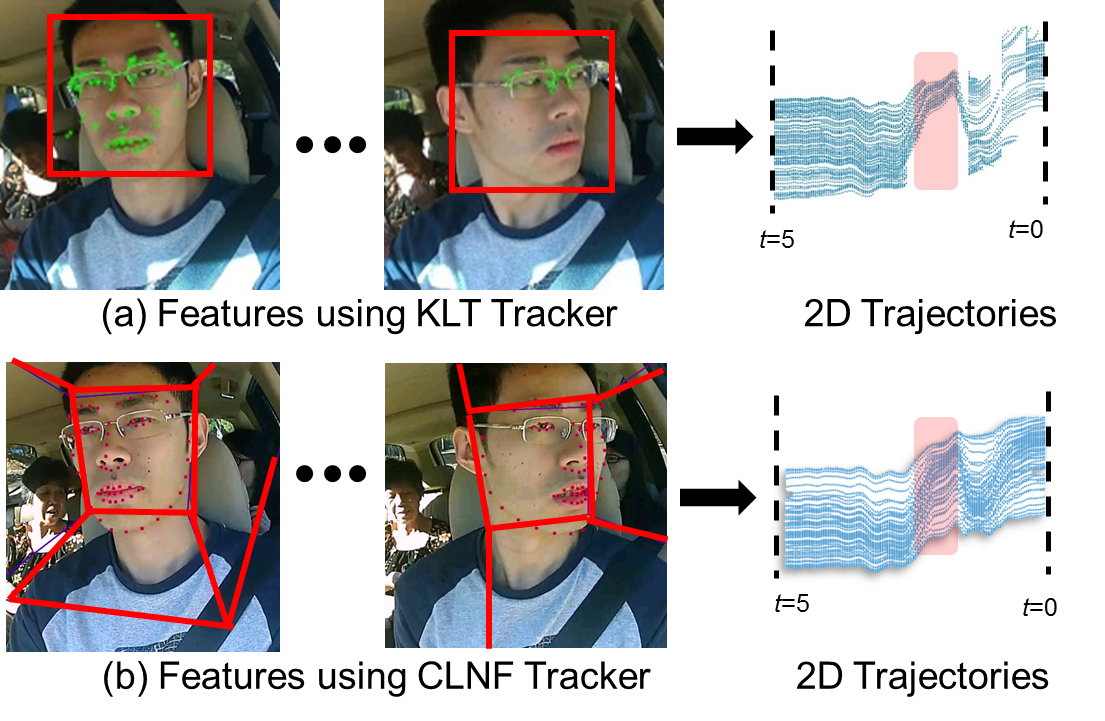}
\vspace{\captionReduceTop}
\caption{\textbf{Improved features for maneuver anticipation.} We track facial landmark points using the CLNF tracker~\citep{Baltrusaitis13} which results in more consistent 2D trajectories as compared to the KLT tracker~\citep{Shi94} used by Jain et al.~\citep{Jain15}. Furthermore, the CLNF also gives an estimate of the driver's 3D head pose.}
\vspace{1.5\captionReduceBot}
\label{fig:feature_compare}
\end{figure}

\vspace{\subsectionReduceTop}
\subsection{Features for maneuver anticipation}
\vspace{\subsectionReduceBot}
\label{subsec:features}

In the vision pipeline of Jain et al.~\citep{Jain15}, the driver facing camera detects discriminative points on the driver's face and tracks the detected points across frames using the KLT tracker~\citep{Shi94}. The tracking generates 2D optical flow trajectories in the image plane. From these trajectories the horizontal and angular movements of the face are extracted, and these movements are binned into histogram features for every frame. These histogram features are aggregated over 20 frames (i.e. 0.8 seconds of driving) and constitute the feature vector  $\ve{x}_t \in \mathbb{R}^9$. 
Further context for anticipation comes from the camera facing the road, the GPS, 
and the vehicle's dynamics. This is denoted by the feature vector $\ve{z}_t \in \mathbb{R}^6$, and includes the lane information, the road artifacts such as intersections, and the vehicle's speed. 
We refer the reader to Jain et al.~\citep{Jain15} for more information on their features.

\vspace{\subsectionReduceTop}
\subsection{3D head pose and facial landmark features}
\vspace{\subsectionReduceBot}
\label{subsec:3dpose}

We now propose new features for maneuver anticipation which significantly improve upon the features from Jain et al.~\citep{Jain15}. Instead of tracking discriminative points on the driver's face we use the Constrained Local Neural Field (CLNF) model~\citep{Baltrusaitis13} and track 68 fixed  landmark points on the driver's face. CLNF is particularly well suited for driving scenarios due its ability to handle a wide range of head pose and illumination variations. As shown in Figure~\ref{fig:feature_compare} CLNF offers us two distinct benefits over the features from Jain et al.~\citep{Jain15} (i) while discriminative facial points may change from situation to situation, tracking fixed landmarks results in consistent optical flow trajectories which adds to robustness; and (ii) CLNF also allows us to estimate the 3D head pose of the driver's face by minimizing  error in the projection of a generic 3D mesh model of the face w.r.t. the 2D location of landmarks in the image. The histogram features generated from the optical flow trajectories along with the 3D head pose features (yaw, pitch and row), give us $\ve{x}_t \in \mathbb{R}^{12}$.

In Section~\ref{experiments} we present results with the features from Jain et al.~\citep{Jain15}, as well as the results with our improved features obtained from the CLNF model.


\vspace{\sectionReduceTop}
\section{Experiments}
\vspace{\sectionReduceBot}
\label{experiments}
We evaluate our proposed architecture on the task of maneuver anticipation~\citep{BoschURBAN,Jain15,Morris11,Tawari14b}. 
This is an important problem because in the US alone 33,000 people die in road accidents every year -- a majority of which are due to dangerous maneuvers. Advanced Driver Assistance Systems (ADAS) have made driving safer by alerting drivers whenever they commit a dangerous maneuver. Unfortunately, many accidents are unavoidable because by the time drivers are alerted it is already too late. Maneuver anticipation can avert many accidents by alerting drivers \textit{before} they perform a dangerous maneuver~\citep{Rueda04}.

We evaluate on the driving data set publicly released by Jain et al.~\citep{Jain15}. The data set consists of 1180 miles of natural freeway and city driving collected from 10 drivers over a period of two months. It contains videos with both inside and outside views of the vehicle, the vehicle's speed, and GPS coordinates. The data set is annotated with 700 events consisting of 274 lane changes, 131 turns, and 295 randomly sampled instances of driving straight. Each lane change and turn is also annotated with the start time of the maneuver, i.e. right before the wheel touches the lane marking or the vehicle yaws at the intersection, respectively. We augment the data set using the technique described in Section~\ref{subsec:augmentation} and generate 2250 events from the 700 original events. We train our deep learning architectures on the augmented data. {We will make our code for data augmentation and maneuver anticipation publicly available at: \url{http://www.brain4cars.com}}

We compare our deep RNN architecture with the following baseline algorithms:
\begin{packed_enum}
\item \textit{Chance:} Uniformly randomly anticipates a maneuver.
\item \textit{Random-forest:} A discriminative classifier that learns an ensemble of 150 decision trees.
\item \textit{SVM~\citep{Morris11}:} Support Vector Machine classifier used by Morris et al.~\citep{Morris11} for maneuver anticipation. 
\item \textit{IOHMM~\citep{Jain15}:} Input-Output Hidden Markov Model~\citep{Bengio95} used by Jain et al.~\citep{Jain15} for maneuver anticipation.  
\item \textit{AIO-HMM~\citep{Jain15}:} This model extends IOHMM by including autoregressive connections in the output layer. AIO-HMM achieved state-of-the-art performance in Jain et al.~\citep{Jain15}. 
\end{packed_enum}

In order to study the effect of our design choices we also compare the following modifications of our architecture:

\begin{packed_enum}
\setcounter{enumi}{5}
\item \textit{Simple-RNN (S-RNN):} In this architecture sensor streams are fused by simple concatenation and then passed through a single RNN with LSTM units. 
\item \textit{Fusion-RNN-Uniform-Loss (F-RNN-UL):} In this architecture sensor streams are passed through separate RNNs, and the high-level representations from RNNs are then fused via a fully-connected layer. The loss at each time step takes the form $-\log(y_t^k)$. 
\item \textit{Fusion-RNN-Exp-Loss (F-RNN-EL):} This architecture is similar to F-RNN-UL, except that the loss exponentially grows with time $-e^{-(T-t)}\log(y_t^k)$. 
\end{packed_enum}

We use the RNN and LSTM implementations provided by Jain~\citep{Neuralmodels}. In our RNNs we use a single layer LSTM of size 64 with sigmoid gate activations and tanh activation for hidden representation. Our fully connected fusion layer uses tanh activation and outputs a 64 dimensional vector. Our overall architecture (F-RNN-EL and F-RNN-UL) have nearly 25,000 parameters that are learned using RMSprop~\citep{Dauphin15}. 

\vspace{\subsectionReduceTop}
\subsection{Evaluation setup}
\vspace{\subsectionReduceBot}
 
We follow an evaluation setup similar to Jain et al.~\citep{Jain15}. Algorithm~\ref{alg:inference} shows the inference steps for maneuver anticipation. At each time step $t$, features $\ve{x}_t$ and $\ve{z}_t$ are computed over the last 0.8 seconds of driving (20 frames). Using the temporal context $\{(\ve{x}_1,...,\ve{x}_t),\; (\ve{z}_1,...,\ve{z}_t)\}$, each anticipation algorithm computes the probability $\ve{y}_t$ for maneuvers in $ \mathcal{M} =$ \{\textit{left lane change, right lane change, left turn, right turn, driving straight}\}. The prediction threshold is denoted by $p_{th}\in(0,1]$ in Algorithm~\ref{alg:inference}. The algorithm predicts \textit{driving straight} if none of the softmax probabilities for the other maneuvers exceeds $p_{th}$.


In order to evaluate an anticipation algorithm, we compute the following quantities for each maneuver $m \in \mathcal{M}$: (i) $N_m$: the total number of instances of maneuver $m$; (ii) $TP_m$: the number of instances of maneuver $m$ correctly predicted by the algorithm; and (iii) $P_m$: the number of times the algorithm predicts $m$. Based on these quantities we evaluate the  precision and recall of an anticipation algorithm as defined below: 

\vspace{-0.15in}
{\small \begin{align}
\label{eq:precision} Pr &= \frac{1}{|M|-1}\sum_{m \in \mathcal{M} \backslash \{\textit{driving straight}\}}\left(\frac{TP_m}{P_m}\right)\\
\label{eq:recall} Re &= \frac{1}{|M|-1}\sum_{m \in \mathcal{M} \backslash \{\textit{driving straight}\}}\left(\frac{TP_m}{N_m}\right)
\end{align}  }
We should note that  \textit{driving straight} maneuver is not included in evaluating  precision~\eqref{eq:precision} and recall~\eqref{eq:recall}. This is because anticipation algorithms by default predict \textit{driving straight} when they are not confident about other maneuvers.  For each anticipation algorithm, we choose a prediction threshold  $p_{th}$  that maximizes their $F1$ score: $F1 = 2*Pr*Re/(Pr+Re)$. In addition to precision and recall, we also measure the interval between the time an algorithm makes a prediction and the start of maneuver. We refer to this as the \textit{time-to-maneuver} and  denote it with $t_{before}$ in Algorithm~\ref{alg:inference}. We uniformly randomly partition the data set into five folds and report  results using 5-fold cross-validation. We train on four folds and test on the fifth fold, and report the metrics averaged over the five folds. 

\begin{table*}[t!]
\centering
\caption{{\textbf{Maneuver Anticipation Results.} Average \textit{precision}, \textit{recall} and \textit{time-to-maneuver} are computed from 5-fold cross-validation. Standard error is also shown. Algorithms are compared on the features from Jain et al.~\citep{Jain15}.}}
\vspace{0.2\captionReduceBot}
\resizebox{1\textwidth}{!}{
\centering
\begin{tabular}{cr|ccc|ccc|ccc}
&  &\multicolumn{3}{c}{Lane change}&\multicolumn{3}{|c}{Turns}&\multicolumn{3}{|c}{All maneuvers}\\
\cline{1-11}
\multicolumn{2}{c|}{\multirow{2}{*}{Method}} & \multirow{2}{*}{$Pr$ (\%)}  & \multirow{2}{*}{$Re$ (\%)} & Time-to-  & \multirow{2}{*}{$Pr$ (\%)} & \multirow{2}{*}{$Re$ (\%)} & Time-to-  & \multirow{2}{*}{$Pr$ (\%)} & \multirow{2}{*}{$Re$ (\%)}  & Time-to- \\ 
& & & &  maneuver (s) &  & &  maneuver (s) &  & & maneuver (s)\\\hline
&Chance	&	33.3		&	33.3		&	-	&	33.3		&	33.3		&	-	&	20.0		&	20.0		&	-\\
&SVM~\citep{Morris11}	&	73.7 $\pm$ 3.4	&	57.8 $\pm$ 2.8	&	2.40 		&	64.7 $\pm$ 6.5	&	47.2 $\pm$ 7.6	&	2.40 		&	43.7 $\pm$ 2.4	&	37.7 $\pm$ 1.8	& 1.20\\
&Random-Forest &   71.2 $\pm$ 2.4  &   53.4 $\pm$ 3.2  &   3.00    &   68.6 $\pm$ 3.5  & 44.4 $\pm$ 3.5  &   1.20      &   51.9 $\pm$ 1.6 &   27.7 $\pm$ 1.1  & 1.20\\
&IOHMM~\citep{Jain15}		&	81.6 $\pm$ 1.0	&	{79.6 $\pm$ 1.9}	&	3.98  	&	77.6 $\pm$ 3.3	&		{75.9 $\pm$ 2.5}	&	4.42 		&	74.2 $\pm$ 1.7	&	71.2 $\pm$ 1.6	&	3.83 \\
&AIO-HMM~\citep{Jain15}		&	{83.8 $\pm$ 1.3}	&	79.2 $\pm$ 2.9	&	3.80  		&	{80.8	 $\pm$ 3.4}	&		75.2 $\pm$ 2.4	&	4.16 		&	{77.4 $\pm$ 2.3}	&	{71.2 $\pm$ 1.3}	&	3.53 \\\hline
 & S-RNN & 85.4 $\pm$ 0.7 & 86.0 $\pm$ 1.4 & 3.53 & 75.2 $\pm$ 1.4 & 75.3 $\pm$ 2.1 & 3.68 & 78.0 $\pm$ 1.5 & 71.1 $\pm$ 1.0 & 3.15 \\
\textit{Our}&F-RNN-UL & \textbf{92.7} $\pm$ 2.1 & 84.4 $\pm$ 2.8 & 3.46 & 81.2 $\pm$ 3.5 & 78.6 $\pm$ 2.8 & 3.94 & 82.2 $\pm$ 1.0 & 75.9 $\pm$ 1.5 & 3.75 \\
\textit{Methods}&F-RNN-EL & 88.2 $\pm$ 1.4 & \textbf{86.0} $\pm$ 0.7 & 3.42 & \textbf{83.8} $\pm$ 2.1 & \textbf{79.9} $\pm$ 3.5 & 3.78 & \textbf{84.5} $\pm$ 1.0 & \textbf{77.1} $\pm$ 1.3 & 3.58\\
\end{tabular}
}
\label{tab:prscore}
\vspace{-0.25in}
\end{table*}

\begin{algorithm}[t]\caption{Maneuver anticipation}
\begin{algorithmic}
\STATE \textbf{Initialize} $m^* = \textit{driving straight}$\\
\STATE \textbf{Input} Features $\{(\ve{x}_1,...,\ve{x}_T),(\ve{z}_1,...,\ve{z}_T)\}$ and prediction threshold $p_{th}$
\STATE \textbf{Output} Predicted maneuver $m^*$
\WHILE{$t=1$ to $T$}
        \STATE Observe features $(\ve{x}_1,...,\ve{x}_t)$ and $(\ve{z}_1,...,\ve{z}_t)$
        \STATE Estimate probability $ \ve{y}_t$ of each maneuver in $\mathcal{M}$
        \STATE $m_t^*=\argmax_{m\in\mathcal{M}}\ve{y}_t$
		\IF{$m_t^* \neq \textit{driving straight}$ \& $\ve{y}_t\{m_t^*\} > p_{th} $}        
			\STATE $m^* = m_t^*$
			\STATE $t_{before} = T - t$
			\STATE \textbf{break}
		\ENDIF

\ENDWHILE
\STATE \textbf{Return} $m^*,t_{before}$
\end{algorithmic}
\label{alg:inference}
\end{algorithm}

\vspace{\subsectionReduceTop}
\subsection{Results}
\vspace{\subsectionReduceBot}

We evaluate anticipation algorithms on the maneuvers not seen during training with the following three prediction settings: (i) Lane change: algorithms only anticipate lane changes, i.e. $ \mathcal{M} =$ \{\textit{left lane change, right lane change, driving straight}\}. This setting is relevant for freeway driving; (ii) Turns: algorithms only anticipate turns, i.e. $ \mathcal{M} =$ \{\textit{left turn, right turn, driving straight}\}; and (iii) All maneuvers: algorithms anticipate all five maneuvers. Among these prediction settings, predicting all five maneuvers is the hardest. 

Table~\ref{tab:prscore} compares the performance of the baseline anticipation algorithms and the variants of our deep learning model. All algorithms in Table~\ref{tab:prscore} were evaluated on the features provided by Jain et al.~\citep{Jain15}, which ensures a fair comparison. We observe that variants of our architecture outperform the previous state-of-the-art a majority of the time. This improvement in performance is because RNNs with LSTM units are very expressive models, and unlike Jain et al.~\citep{Jain15} they do not make any assumption about the generative nature of the problem.

The performance of several variants of our architecture, reported in Table~\ref{tab:prscore},  justifies our design decisions to reach the final architecture as discussed here.
S-RNN performs a very simple fusion by concatenating the two sensor streams. On the other hand, F-RNN models each sensor stream with a separate RNN and then uses a fully connected layer at each time step to fuse the high-level representations. This form of sensory fusion is  more principled since the sensor streams represent different data modalities.  Fusing high-level representations instead of concatenating raw features gives a significant improvement in performance, as shown in Table~\ref{tab:prscore}.  When predicting all maneuvers, F-RNN-EL has a 6\%  higher precision and recall than S-RNN.

As shown in Table~\ref{tab:prscore}, exponentially growing the loss improves performance. Our new loss scheme  penalizes the network proportional to the length of context it has seen.  When predicting all maneuvers, we observe that F-RNN-EL shows an improvement of 2\% in precision and recall over F-RNN-UL. We conjecture that exponentially growing the loss acts like a regularizer. It reduces the risk of our network over-fitting early in time when there is not enough context available. Furthermore, the time-to-maneuver remains comparable for F-RNN  with and without exponential loss.

We study the effect of our improved features in Table~\ref{tab:features}. We replace the pipeline for extracting features from the driver's face~\citep{Jain15} by a Constrained Local Neural Field (CLNF) model~\citep{Baltrusaitis13}. Our new vision pipeline tracks 68 facial landmark points and estimates the driver's 3D head pose as described in Section~\ref{subsec:features}. 
We see a significant, 6\% increase in precision and 10\% increase in recall of F-RNN-EL when using features from our new vision pipeline. This increase in performance is attributed to the following reasons: (i) robustness of CLNF model to variations in illumination and head pose; (ii) 3D head-pose features are very informative for understanding the driver's intention; and (iii) optical flow trajectories generated by tracking facial landmark points represent head movements better, as shown in Figure~\ref{fig:feature_compare}.

\begin{table}[h]
\centering
\vspace{.05in}
\caption{\textbf{3D head-pose features.} In this table we study the effect of better features with best performing algorithm from Table~\ref{tab:prscore} in `All maneuvers' setting. We use~\citep{Baltrusaitis13} to track 68 facial landmark points and estimate 3D head-pose.}
\vspace{0.5\captionReduceBot}
{
\newcolumntype{P}[2]{>{\footnotesize#1\hspace{0pt}\arraybackslash}p{#2}}
\setlength{\tabcolsep}{2pt}
\centering
\resizebox{\hsize}{!}{
\begin{tabular}
{@{}p{0.40\linewidth}| P{\centering}{16mm}P{\centering}{16mm}P{\centering}{16mm}@{}}
\multirow{2}{*}{Method} & \multirow{2}{*}{$Pr$ (\%)}  & \multirow{2}{*}{$Re$ (\%)} & Time-to-  \\ 
 & & &  maneuver (s)\\\hline
F-RNN-EL &  {84.5} $\pm$ 1.0 & {77.1} $\pm$ 1.3 & 3.58\\
F-RNN-EL w/ 3D head-pose &  \textbf{90.5} $\pm$ 1.0 & \textbf{87.4} $\pm$ 0.5 & 3.16\\
\end{tabular}
}}
\label{tab:features}
\end{table}

\begin{figure*}[t]
\centering
\begin{subfigure}[b]{.23\textwidth}
	\includegraphics[width=\linewidth]{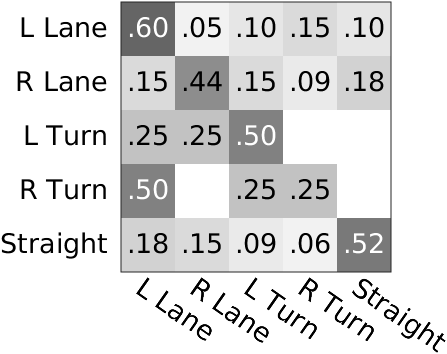}
	\vspace{2.2\captionReduceTop}
	\caption{SVM~\citep{Morris11}}
	\end{subfigure}
\begin{subfigure}[b]{.23\textwidth}
	\includegraphics[width=\linewidth]{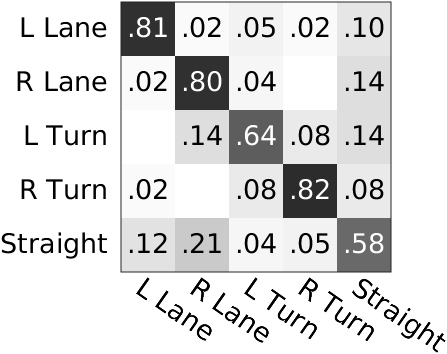}
	\vspace{2.2\captionReduceTop}
	\caption{AIO-HMM~\citep{Jain15}}
	\end{subfigure}	
	\begin{subfigure}[b]{.23\textwidth}
	\includegraphics[width=\linewidth]{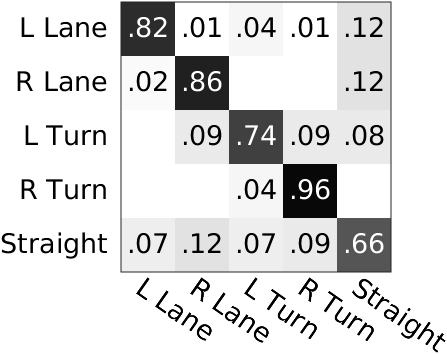}
	\vspace{2.2\captionReduceTop}
	\caption{F-RNN-EL}
	\end{subfigure}	
	\begin{subfigure}[b]{.23\textwidth}
	\includegraphics[width=\linewidth]{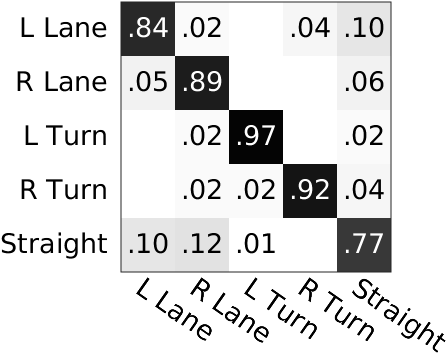}
	\vspace{2.2\captionReduceTop}
	\caption{F-RNN-EL w/ 3D-pose}
	\end{subfigure}
	\vspace{\captionReduceTop}
	\caption{\textbf{Confusion matrix} of different algorithms when jointly predicting all the maneuvers. Predictions made by algorithms are represented by rows and actual maneuvers are represented by columns. Numbers on the diagonal represent precision.}
	\vspace{1.2\captionReduceBot}
	\label{fig:confmat}	
\end{figure*}

\begin{figure}[t]
\centering
\vspace{-0.15in}
\includegraphics[width=0.8\linewidth]{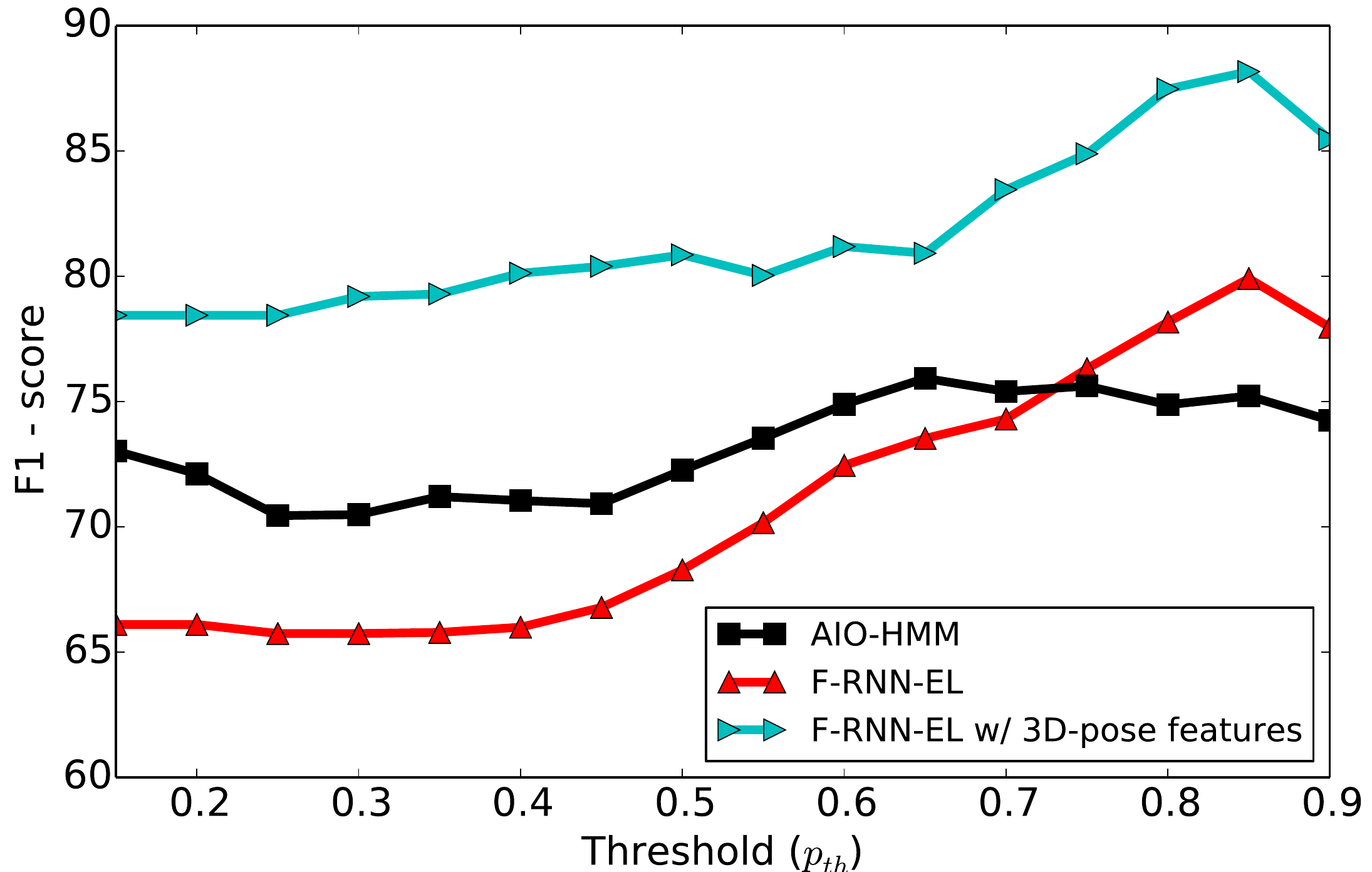}
\vspace{0.8\captionReduceTop}
\caption{\textbf{Effect of prediction threshold $p_{th}$.} At test time an algorithm makes a prediction only when it is at least $p_{th}$ confident in its prediction. This plot shows how F1-score vary with change in prediction threshold.}
\vspace{1.25\captionReduceBot}
\label{fig:f1score}
\end{figure}

The confusion matrix in Figure~\ref{fig:confmat} shows the precision for each maneuver. F-RNN-EL gives a higher precision than AIO-HMM on every maneuver when both algorithms are trained on same features (Fig.~\ref{fig:confmat}c). Our new vision pipeline further improves the precision of F-RNN-EL on all maneuvers (Fig.~\ref{fig:confmat}d). Additionally, both F-RNN and AIO-HMM perform significantly better than previous work on maneuver anticipation by Morris et al.~\citep{Morris11} (Fig.~\ref{fig:confmat}a).

In Figure~\ref{fig:f1score} we study how F1-score varies as we change the prediction threshold $p_{th}$.  We make the following observations: (i) The F1-score does not undergo large variations with changes to the prediction threshold. Hence, it allows practitioners  to fairly trade-off between the precision and recall without hurting the F1-score by much; and (ii)   the maximum F1-score attained by F-RNN-EL is 4\% more than AIO-HMM when compared on the same features and 13\% more with our new vision pipeline.
In {Tables~\ref{tab:prscore} and \ref{tab:features}}, we used the threshold values which gave the highest F1-score.

\vspace{\sectionReduceTop}
\section{Conclusion}
\vspace{\sectionReduceBot}

In this work we addressed the problem of anticipating maneuvers several seconds before they happen. This problem requires the modeling of long temporal dependencies and the fusion of multiple sensory streams. We proposed a novel deep learning architecture based on Recurrent Neural Networks (RNNs) with Long Short-Term Memory (LSTM) units for anticipation. Our architecture learns to fuse multiple sensory streams, and by training it in a sequence-to-sequence prediction manner, it explicitly learns to anticipate using only a partial temporal context. We also proposed a novel loss layer for anticipation which prevents over-fitting.

Our deep learning architecture outperformed the previous state-of-the-art on 1180 miles of natural driving data set. It improved the precision from 78\% to 84.5\% and recall from 71.1\% to 77.1\%. We further showed that {improving head tracking and} including the driver's 3D head pose as a feature gives a significant boost in performance by increasing the precision to 90.5\% and recall to 87.4\%. We believe that our approach is widely applicable to many activity anticipation problems. As more anticipation data sets become publicly available, we expect to see a similar improvement in performance with our architecture.
\\~\\~
\textbf{Acknowledgement.} We thank NVIDIA for the donation of a K40 GPU used in
this research. We also thank Silvio Savarese for useful discussions. This work
was supported by National Robotics Initiative (NRI) award 1426452, Office of Naval
Research (ONR) award N00014-14-1-0156, and by Microsoft Faculty Fellowship and NSF Career
Award to one of us (Saxena).\vspace{2\sectionReduceBot}

\end{document}